\documentclass[letterpaper, 10 pt, conference]{ieeeconf}  

\IEEEoverridecommandlockouts                              

\usepackage{graphicx}
\usepackage{microtype}

\usepackage{tikz, pgfplots, pgfplotstable}
\usetikzlibrary{calc}
\usetikzlibrary{positioning}
\usepgflibrary{fpu}

\usetikzlibrary{external}
\pgfplotsset{compat=newest}

\usepackage{amsmath}

\usepackage{subcaption}
\DeclareCaptionLabelSeparator{periodspace}{.\quad}
\usepackage{algorithm, algorithmic}
\floatstyle{plaintop}
\restylefloat{algorithm}

\usepackage{placeins}
\usepackage{url}

\usepackage{amsmath}

\def\addlegendimage{\csname pgfplots@addlegendimage\endcsname}

\usepackage{booktabs}

\title{\LARGE \bf Data augmentation with Symbolic-to-Real Image Translation GANs\\ for Traffic Sign Recognition}

\author{Nour Soufi$^{1}$
    \thanks{$^{1}$ Nour Soufi is with the Hochschule Bonn-Rhein-Sieg and Fraunhofer IAIS, Sank Augustin, Germany
    {\tt\small noursufi@gmail.com}}%
\and
Matias Valdenegro-Toro$^{2}$
    \thanks{$^{2}$ Matias Valdenegro-Toro is with the German Research Center for Artificial Intelligence, Robotics Innovation Center. Robert-Hooke-Strasse 1, 28359, Bremen, Germany.
    {\tt\small matias.valdenegro@dfki.de}}%
}

\begin{document}

\maketitle
\thispagestyle{empty}
\pagestyle{empty}

\begin{abstract}
    Traffic sign recognition is an important component of many advanced driving assistance systems, and it is required for full autonomous driving. Computational performance is usually the bottleneck in using large scale neural networks for this purpose. SqueezeNet \cite{SqueezeNet} is a good candidate for efficient image classification of traffic signs, but in our experiments it does not reach high accuracy, and we believe this is due to lack of data, requiring data augmentation.
	Generative adversarial networks can learn the high dimensional distribution of  empirical data, allowing the generation of new data points. In this paper we apply pix2pix GANs \cite{pix2pix2016} architecture to generate new traffic sign images and evaluate the use of these images in data augmentation. We were motivated to use pix2pix to translate symbolic sign images to real ones due to the mode collapse in Conditional GANs.
    Through our experiments we found that data augmentation using GAN can increase classification accuracy for circular traffic signs from 92.1\% to 94.0\%, and for triangular traffic signs from 93.8\% to 95.3\%, producing an overall improvement of 2\%. However some traditional augmentation techniques can outperform GAN data augmentation, for example contrast variation in circular traffic signs (95.5\%) and displacement on triangular traffic signs (96.7 \%). Our negative results shows that while GANs can be naively used for data augmentation, they are not always the best choice, depending on the problem and variability in the data.
\end{abstract}

	\section{Introduction}
\label{sec:introduction}

\noindent Traffic sign recognition is an integral part of Advanced Driving Assistance Systems. It consists of two principal sub-problems: Detection and Classification. Convolutional Neural Networks (CNNs) are one of the most promising methods to solve such problems \cite{7605450}. A common issue with this kind of models is the large training sets required to achieve good generalization \cite{8001185}, which motivates the use of data augmentation, specially when using real-world imbalanced and low-sample datasets.

The development of data augmentation techniques is problem, domain, and data dependent. There have been some efforts to automate this process by learning augmentations directly from data. A good choice for this problem is to use a generative model that can learn to generate new novel samples.

In this work we study how generative adversarial networks (GANs) can be used to generate complicated images like traffic signs, and we analyze how this can be applied to perform data augmentation and enlarge the size of the training set. We adjust the implementation of an image-to-image translation model (pix2pix), which is a special application of GANs, to map from a symbolic traffic sign image to a real image. This is motivated by the failure of a vanilla GAN to produce varied samples on our dataset.

This paper is organized as follows. First we describe previous work related to data augmentation and traffic sign recognition. Then we give an introduction about GANs, which is the main focus in this work, after that we explain the methods that we used, finally we introduce  our results and conclusion. 

\section{State of The Art}

Traffic sign classification previous works falls into two main sections detection and classification. In this work we focus on the classification part, where the task is to classify the detected sign under one of traffic signs sub-categories like speed limit, no stopping, priority road ...etc. Different machine learning techniques were applied to solve this task, multi scale CNN \cite{6033589}, committee of CNNs \cite{6033458}, SVM \cite{6033395}  and random forests \cite{6033494}. In general CNNs achieved high classification accuracy, more than 99\%, but the problem is still the computation time that CNNs requires.\\	In our solution we applied a special architecture of CNN SqueezeNet \cite{SqueezeNet}  for traffic sign classification, which was proven to be faster than basic CNN, but it  gave less classification accuracy, for this reason we try to increase the accuracy by applying data augmentation.

Data augmentation or (jittering)  became very important and essential in classification problems to increase the number of samples in training set, especially when CNN is used for classification. Data augmentation produces more images in the training set, which look similar to the images that already exist and belong to the same classes, but still slightly different giving classification network more generalization \cite{6766231}.

Most popular augmentation methods for image datasets depend on geometrical transformations; like rotation \cite{8279326}, scaling \cite{6766231}, and displacement \cite{ashiquzzaman2017applying}, or color transformations; like ZCA-whitening \cite{ashiquzzaman2017applying}, PCA (Principle component analysis) \cite{NIPS2012_4824}, color casting, vignetting and lens distortion \cite{wu2015deep}. In this work we compare some of the traditional methods to GANs augmentation.

GANs were proposed by Goodfellow et al. \cite{NIPS2014_5423}, they can be described as a pair of neural networks that are training simultaneously and competing with each other, simulating a min-max game. The first network is called the generator, which learns to generate synthetic images that look similar to the real ones from a dataset. The second network, the discriminator, tries to distinguish real images from synthetic ones.	There were also many improvements and variations from basic GANs, one of them is Conditional GAN which  was introduced in \cite{mirza2014conditional}. In this type both the generator and discriminator receive a special condition as an extra input $y$. This condition could be a class index or any other conditional information.

Pix2pix \cite{pix2pix2016} was presented as an application of conditional GANs, where the authors apply image to image translation. By using a suitable dataset, their approach can learn the solution for many problems, converting gray scale image to RGB image, creating photos from edges, converting labels to street scenes, and others. As we mentioned Conditional GAN receives an extra input as a condition, in Pix2pix this condition is the input image, whereas they also eliminate the noise input vector, as they have seen in their experiments that the generator ignores noise input while training.

\begin{figure} [htb!]	
	\centering
	\raisebox{-\height}{\includegraphics[width=8cm]{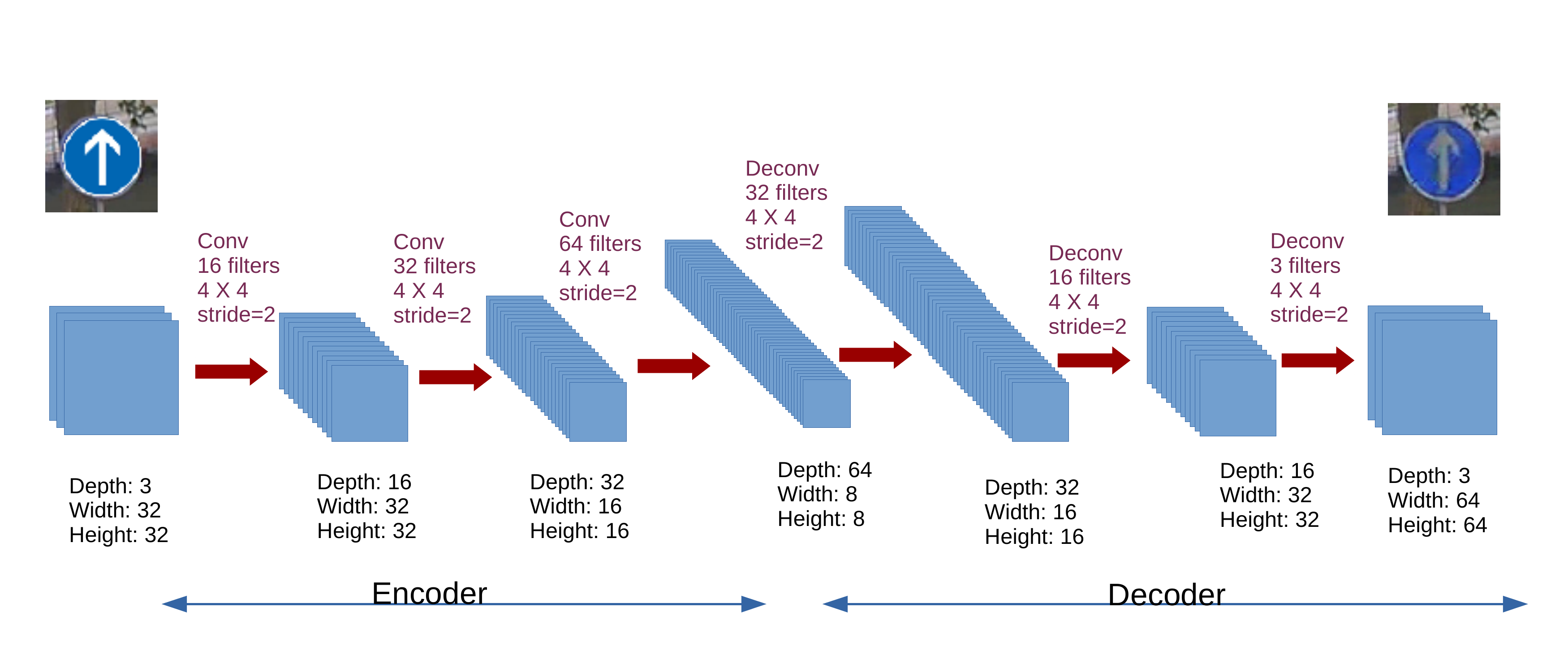}}
	\caption{Pix2pix Generator Architecture}
	\label{fig:Generator}
\end{figure}

\section{Traffic Sign Augmentation with pix2pix}

\begin{figure} [htb!]
	\centering
	\raisebox{-\height}{\includegraphics[width=8cm]{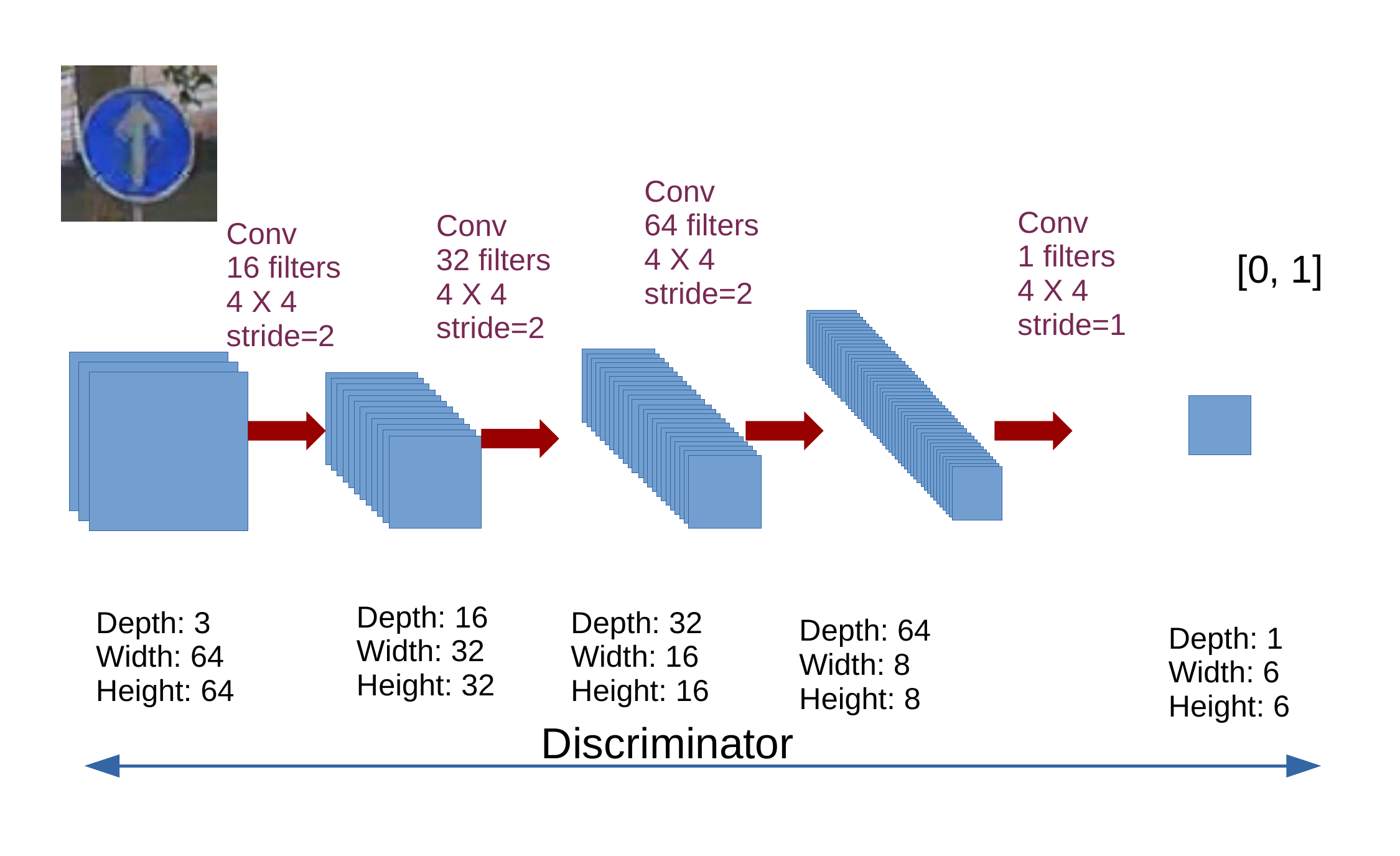}}
	\caption{Pix2pix Discriminator Architecture}
	\label{fig:Discriminator}
\end{figure}

In this architecture the generator has an encoder-decoder setting. In the encoder the input is down-sampled until the last layer in the encoder (bottleneck layer), then it is up sampled in the decoder. By this the output and input of the generator has the same dimensions.
The discriminator architecture  was in (PatchGAN), which means that the input image of the discriminator is divided into patches, and each patch is determined separately whether it is real or fake (generated by the generator). The final decision would be the average of all the patches outputs. We can see in figure \ref{fig:Discriminator} the discriminator architecture.

We propose to perform data augmentation for traffic signs by taking data in the training set, rendering a symbolic traffic sign on top of the bounding box of the real training image, and using pix2pix to transform this image into a real traffic sign. This is motivated by our initial experiments with a vanilla GAN, where we found that not enough variability was being generated by the GAN. This meant that for many noise vectors $z$ the GAN generated the same image, implying mode collapse. We hypothesize that this is due to our small dataset of traffic signs and the low intra-class variability natural to the design of a traffic sign.

\begin{figure*}[htb!]
   \centering
   \begin{subfigure}{0.15\textwidth}
       \centering
       \raisebox{-\height}{\includegraphics[width=2cm]{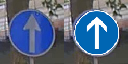}}\hfill
       \caption*{Class 0}
   \end{subfigure}
   \begin{subfigure}{0.15\textwidth}
       \centering
       \raisebox{-\height}{\includegraphics[width=2cm]{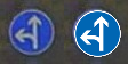}}\hfill
       \caption*{Class 1}
   \end{subfigure}
   \begin{subfigure}{0.15\textwidth}
       \centering
       \raisebox{-\height}{\includegraphics[width=2cm]{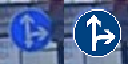}}\hfill
       \caption*{Class 2}
   \end{subfigure}
   \begin{subfigure}{0.15\textwidth}
       \centering
       \raisebox{-\height}{\includegraphics[width=2cm]{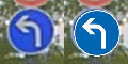}}\hfill
       \caption*{Class 3}
   \end{subfigure}
   \begin{subfigure}{0.15\textwidth}
       \centering
       \raisebox{-\height}{\includegraphics[width=2cm]{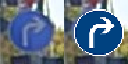}}\hfill
       \caption*{Class 4}
   \end{subfigure}
   \begin{subfigure}{0.15\textwidth}
       \centering
       \raisebox{-\height}{\includegraphics[width=2cm]{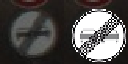}}\hfill
       \caption*{Class 5}
   \end{subfigure}
   \\
   \begin{subfigure}{0.15\textwidth}
       \centering
       \raisebox{-\height}{\includegraphics[width=2cm]{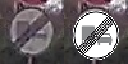}}\hfill
       \caption*{Class 6}
   \end{subfigure}
   \begin{subfigure}{0.15\textwidth}
       \centering
       \raisebox{-\height}{\includegraphics[width=2cm]{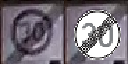}}\hfill
       \caption*{Class 7}
   \end{subfigure}
   \begin{subfigure}{0.15\textwidth}
       \centering
       \raisebox{-\height}{\includegraphics[width=2cm]{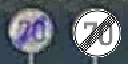}}\hfill
       \caption*{Class 8}
   \end{subfigure}
   \begin{subfigure}{0.15\textwidth}
       \centering
       \raisebox{-\height}{\includegraphics[width=2cm]{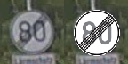}}\hfill
       \caption*{Class 9}
   \end{subfigure}
   \begin{subfigure}{0.15\textwidth}
       \centering
       \raisebox{-\height}{\includegraphics[width=2cm]{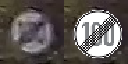}}\hfill
       \caption*{Class 10}
   \end{subfigure}
   \begin{subfigure}{0.15\textwidth}
       \centering
       \raisebox{-\height}{\includegraphics[width=2cm]{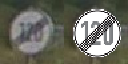}}\hfill
       \caption*{Class 11}
   \end{subfigure}
   \\
   \begin{subfigure}{0.15\textwidth}
       \centering
       \raisebox{-\height}{\includegraphics[width=2cm]{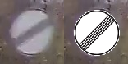}}\hfill
       \caption*{Class 12}
   \end{subfigure}
   \begin{subfigure}{0.15\textwidth}
       \centering
       \raisebox{-\height}{\includegraphics[width=2cm]{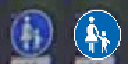}}\hfill
       \caption*{Class 13}
   \end{subfigure}
   \begin{subfigure}{0.15\textwidth}
       \raisebox{-\height}{\includegraphics[width=2cm]{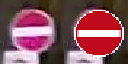}}\hfill
       \caption*{Class 14}
   \end{subfigure}
   \begin{subfigure}{0.15\textwidth}
       \centering
       \raisebox{-\height}{\includegraphics[width=2cm]{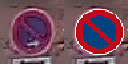}}\hfill
       \caption*{Class 15}
   \end{subfigure}
   \begin{subfigure}{0.15\textwidth}
       \centering
       \raisebox{-\height}{\includegraphics[width=2cm]{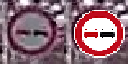}}\hfill
       \caption*{Class 16}
   \end{subfigure}
   \begin{subfigure}{0.15\textwidth}
       \centering
       \raisebox{-\height}{\includegraphics[width=2cm]{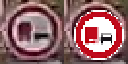}}\hfill
       \caption*{Class 17}
   \end{subfigure}
   \\
   \begin{subfigure}{0.15\textwidth}
       \centering
       \raisebox{-\height}{\includegraphics[width=2cm]{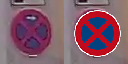}}\hfill
       \caption*{Class 18}
   \end{subfigure}
   \begin{subfigure}{0.15\textwidth}
       \centering
       \raisebox{-\height}{\includegraphics[width=2cm]{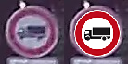}}\hfill
       \caption*{Class 19}
   \end{subfigure}
   \begin{subfigure}{0.15\textwidth}
       \centering
       \raisebox{-\height}{\includegraphics[width=2cm]{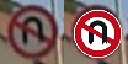}}\hfill
       \caption*{Class 20}
   \end{subfigure}
   \begin{subfigure}{0.15\textwidth}
       \centering
       \raisebox{-\height}{\includegraphics[width=2cm]{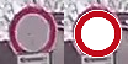}}\hfill
       \caption*{Class 21}
   \end{subfigure}
   \begin{subfigure}{0.15\textwidth}
       \centering
       \raisebox{-\height}{\includegraphics[width=2cm]{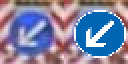}}\hfill
       \caption*{Class 22}
   \end{subfigure}
   \begin{subfigure}{0.15\textwidth}
       \centering
       \raisebox{-\height}{\includegraphics[width=2cm]{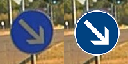}}\hfill
       \caption*{Class 23}
   \end{subfigure}
   \\
   \begin{subfigure}{0.15\textwidth}
       \centering
       \raisebox{-\height}{\includegraphics[width=2cm]{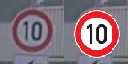}}\hfill
       \caption*{Class 24}
   \end{subfigure}
   \begin{subfigure}{0.15\textwidth}
       \centering
       \raisebox{-\height}{\includegraphics[width=2cm]{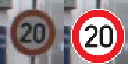}}\hfill
       \caption*{Class 25}
   \end{subfigure}
   \begin{subfigure}{0.15\textwidth}
       \centering
       \raisebox{-\height}{\includegraphics[width=2cm]{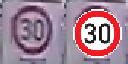}}\hfill
       \caption*{Class 26}
   \end{subfigure}
   \begin{subfigure}{0.15\textwidth}
       \centering
       \raisebox{-\height}{\includegraphics[width=2cm]{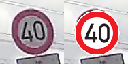}}\hfill
       \caption*{Class 27}
   \end{subfigure}
   \begin{subfigure}{0.15\textwidth}
       \raisebox{-\height}{\includegraphics[width=2cm]{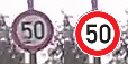}}\hfill
       \caption*{Class 28}
   \end{subfigure}
   \begin{subfigure}{0.15\textwidth}
       \centering
       \raisebox{-\height}{\includegraphics[width=2cm]{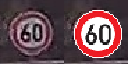}}\hfill
       \caption*{Class 29}
   \end{subfigure}
   \\
   \begin{subfigure}{0.15\textwidth}
       \centering
       \raisebox{-\height}{\includegraphics[width=2cm]{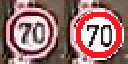}}\hfill
       \caption*{Class 30}
   \end{subfigure}
   \begin{subfigure}{0.15\textwidth}
       \centering
       \raisebox{-\height}{\includegraphics[width=2cm]{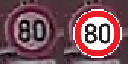}}\hfill
       \caption*{Class 31}
   \end{subfigure}
   \begin{subfigure}{0.15\textwidth}
       \centering
       \raisebox{-\height}{\includegraphics[width=2cm]{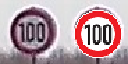}}\hfill
       \caption*{Class 32}
   \end{subfigure}
   \begin{subfigure}{0.15\textwidth}
       \centering
       \raisebox{-\height}{\includegraphics[width=2cm]{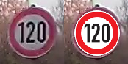}}\hfill
       \caption*{Class 33}
   \end{subfigure}
   \begin{subfigure}{0.15\textwidth}
       \centering
       \raisebox{-\height}{\includegraphics[width=2cm]{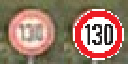}}\hfill
       \caption*{Class 34}
   \end{subfigure}
   \begin{subfigure}{0.15\textwidth}
       \centering
       \raisebox{-\height}{\includegraphics[width=2cm]{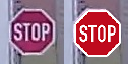}}\hfill
       \caption*{Class 35}
   \end{subfigure}
    \caption{Examples of Pix2pix training set of circular signs, mapping from symbolic to real signs.}
    \label{fig:siblings}
\end{figure*}

\begin{figure*}[htb!]
        \centering
        \begin{subfigure}{0.15\textwidth}
            \centering
            \includegraphics[width=2cm]{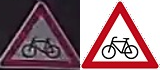}
            \caption*{Class 0}
        \end{subfigure}
        \begin{subfigure}{0.15\textwidth}
            \centering
            \raisebox{-\height}{\includegraphics[width=2cm]{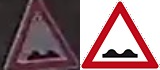}}\hfill
            \caption*{Class 1}
        \end{subfigure}
        \begin{subfigure}{0.15\textwidth}
            \centering
            \raisebox{-\height}{\includegraphics[width=2cm]{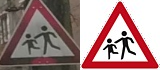}}\hfill
            \caption*{Class 2}
        \end{subfigure}
        \begin{subfigure}{0.15\textwidth}
            \centering
            \raisebox{-\height}{\includegraphics[width=2cm]{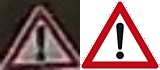}}\hfill
            \caption*{Class 3}
        \end{subfigure}
        \begin{subfigure}{0.15\textwidth}
            \centering
            \raisebox{-\height}{\includegraphics[width=2cm]{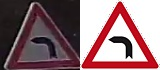}}\hfill
            \caption*{Class 4}
        \end{subfigure}
        \begin{subfigure}{0.15\textwidth}
            \centering
            \raisebox{-\height}{\includegraphics[width=2cm]{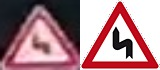}}\hfill
            \caption*{Class 5}
        \end{subfigure}
        \\
        \begin{subfigure}{0.15\textwidth}
            \centering
            \raisebox{-\height}{\includegraphics[width=2cm]{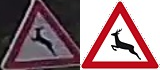}}\hfill
            \caption*{Class 6}
        \end{subfigure}
        \begin{subfigure}{0.15\textwidth}
            \centering
            \raisebox{-\height}{\includegraphics[width=2cm]{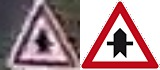}}\hfill
            \caption*{Class 7}
        \end{subfigure}
        \begin{subfigure}{0.15\textwidth}
            \centering
            \raisebox{-\height}{\includegraphics[width=2cm]{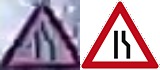}}\hfill
            \caption*{Class 8}
        \end{subfigure}
        \begin{subfigure}{0.15\textwidth}
            \centering
            \raisebox{-\height}{\includegraphics[width=2cm]{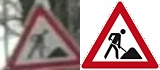}}\hfill
            \caption*{Class 9}
        \end{subfigure}
        \begin{subfigure}{0.15\textwidth}
            \centering
            \raisebox{-\height}{\includegraphics[width=2cm]{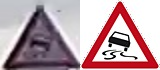}}\hfill
            \caption*{Class 10}
        \end{subfigure}
        \begin{subfigure}{0.15\textwidth}
            \centering
            \raisebox{-\height}{\includegraphics[width=2cm]{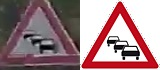}}\hfill
            \caption*{Class 11}
        \end{subfigure}
        \\
        \begin{subfigure}{0.15\textwidth}
            \centering
            \raisebox{-\height}{\includegraphics[width=2cm]{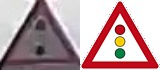}}\hfill
            \caption*{Class 12}
        \end{subfigure}
        \begin{subfigure}{0.15\textwidth}
            \centering
            \raisebox{-\height}{\includegraphics[width=2cm]{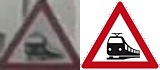}}\hfill
            \caption*{Class 13}
        \end{subfigure}
        \begin{subfigure}{0.15\textwidth}
            \raisebox{-\height}{\includegraphics[width=2cm]{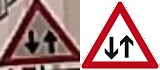}}\hfill
            \caption*{Class 14}
        \end{subfigure}
        \begin{subfigure}{0.15\textwidth}
            \centering
            \raisebox{-\height}{\includegraphics[width=2cm]{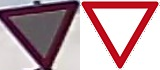}}\hfill
            \caption*{Class 15}
        \end{subfigure}
        \caption{Examples of Pix2pix training set of triangular signs, mapping from symbolic to real signs}\label{fig:siblings_tri}
    \end{figure*}

The pix2pix network is trained with pairs of real and synthetic images of traffic signs, and it learns to map a symbolic traffic sign into a real one, from where new samples can be generated and added to the augmented training set. A sample of training set image pairs is shown in Figure \ref{fig:siblings} for circular signs, and in Figure \ref{fig:siblings_tri} for triangular ones. We tuned the hyper-parameters of two pix2pix instances on circular (Table \ref{tab:accuracyCircular}) and triangular traffic signs (Table \ref{tab:accuracyTriangular}).

\begin{figure*}[htb!]
        
        \begin{center}$
            \begin{array}{ccccccccc}
            \includegraphics[width=1.3cm]{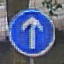}
            &\includegraphics[width=1.3cm]{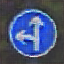}&
            \includegraphics[width=1.3cm]{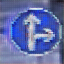}&
            \includegraphics[width=1.3cm]{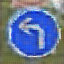}&
            
            \includegraphics[width=1.3cm]{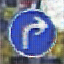}&
            
            \includegraphics[width=1.3cm]{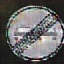}&
            \includegraphics[width=1.3cm]{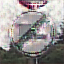}&
            
            \includegraphics[width=1.3cm]{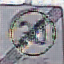}&
            \includegraphics[width=1.3cm]{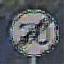}\\
            
            \includegraphics[width=1.3cm]{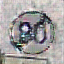}&
            
            \includegraphics[width=1.3cm]{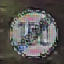}&
            
            \includegraphics[width=1.3cm]{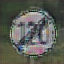}&
            
            \includegraphics[width=1.3cm]{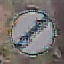}&
            
            \includegraphics[width=1.3cm]{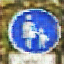}&
            \includegraphics[width=1.3cm]{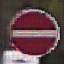}&
            \includegraphics[width=1.3cm]{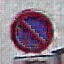}&
            
            \includegraphics[width=1.3cm]{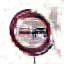}&
            \includegraphics[width=1.3cm]{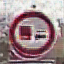}\\
            
            \includegraphics[width=1.3cm]{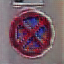}&
            
            \includegraphics[width=1.3cm]{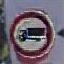}&
            \includegraphics[width=1.3cm]{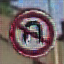}&
            
            \includegraphics[width=1.3cm]{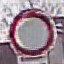}&
            
            \includegraphics[width=1.3cm]{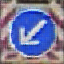}&
            
            \includegraphics[width=1.3cm]{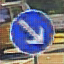}&
            
            \includegraphics[width=1.3cm]{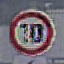}&
            
            \includegraphics[width=1.3cm]{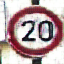}&
            
            \includegraphics[width=1.3cm]{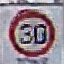}\\
            
            \includegraphics[width=1.3cm]{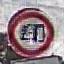}&
            
            \includegraphics[width=1.3cm]{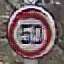}&
            
            \includegraphics[width=1.3cm]{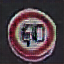}&

            \includegraphics[width=1.3cm]{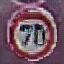}&
            
            \includegraphics[width=1.3cm]{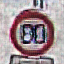}&
            
            \includegraphics[width=1.3cm]{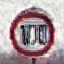}&
            
            \includegraphics[width=1.3cm]{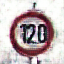}&
            
            \includegraphics[width=1.3cm]{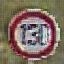}&
            
            \includegraphics[width=1.3cm]{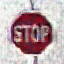}
        \end{array}$
        \end{center}
        \caption{Examples of Circular Traffic Sign Images Generated by Pix2pix}
        \label{fig:output}	
    \end{figure*}

Figure \ref{fig:output} shows some examples of the generated circular traffic sign images. We can notice that for signs with figure inside it (like direction signs, no passing, no U-turn) the generated images looked very similar to real signs. Whereas for signs with digits (end speed limits and speed limits), the figures in the generated images were mostly blurry and unrecognizable.

\section{Evaluation of Traffic Sign Classification}
In this section we evaluate the effect of using GAN-augmented images for traffic sign classification.
    
\textbf{Data}. We possess a dataset containing 61089 training and 30023 testing images of 36 different classes of circular traffic signs (see Figure \ref{fig:siblings}), and another dataset containing 90218 training images 44596 testing images of 16 classes of triangular traffic signs (see Figure \ref{fig:siblings_tri}). These images are rescaled to $64 \times 64$ pixels. From the circular training set, a subset of 5809 high-quality bounding box samples are cropped and used to train the pix2pix GAN, from where we will create new augmented examples by sampling the pix2pix generator given an input symbolic image.

\begin{figure*}[htb!]
            \begin{subfigure}{0.99\textwidth}
                \centering
                \raisebox{-\height}{\includegraphics[width=0.06\textwidth]{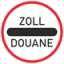}}
                \hfil
            \end{subfigure}
            \begin{subfigure}{0.99\textwidth}
                \raisebox{-\height}{\includegraphics[width=0.06\textwidth]{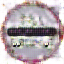}}\hfill
                \raisebox{-\height}{\includegraphics[width=0.06\textwidth]{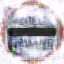}}\hfill
                \raisebox{-\height}{\includegraphics[width=0.06\textwidth]{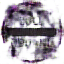}}\hfill
                \raisebox{-\height}{\includegraphics[width=0.06\textwidth]{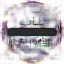}}\hfill
                \raisebox{-\height}{\includegraphics[width=0.06\textwidth]{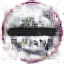}}\hfill
                \raisebox{-\height}{\includegraphics[width=0.06\textwidth]{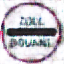}}\hfill
                \raisebox{-\height}{\includegraphics[width=0.06\textwidth]{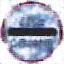}}\hfill
                \raisebox{-\height}{\includegraphics[width=0.06\textwidth]{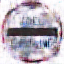}}\hfill
                \raisebox{-\height}{\includegraphics[width=0.06\textwidth]{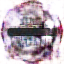}}\hfill
                \raisebox{-\height}{\includegraphics[width=0.06\textwidth]{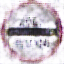}}\hfill                
                \raisebox{-\height}{\includegraphics[width=0.06\textwidth]{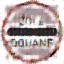}}\hfill
                \raisebox{-\height}{\includegraphics[width=0.06\textwidth]{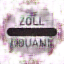}}\hfill
                \raisebox{-\height}{\includegraphics[width=0.06\textwidth]{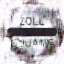}}\hfill
                \raisebox{-\height}{\includegraphics[width=0.06\textwidth]{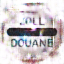}}\hfill
                \raisebox{-\height}{\includegraphics[width=0.06\textwidth]{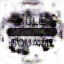}}\hfill
                
                \raisebox{-\height}{\includegraphics[width=0.06\textwidth]{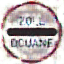}}\hfill
                \raisebox{-\height}{\includegraphics[width=0.06\textwidth]{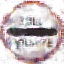}}\hfill
                \raisebox{-\height}{\includegraphics[width=0.06\textwidth]{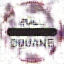}}\hfill
                \raisebox{-\height}{\includegraphics[width=0.06\textwidth]{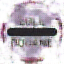}}\hfill
                \raisebox{-\height}{\includegraphics[width=0.06\textwidth]{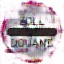}}\hfill                
                \raisebox{-\height}{\includegraphics[width=0.06\textwidth]{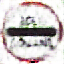}}\hfill
                \raisebox{-\height}{\includegraphics[width=0.06\textwidth]{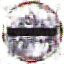}}\hfill
                \raisebox{-\height}{\includegraphics[width=0.06\textwidth]{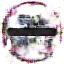}}\hfill
                \raisebox{-\height}{\includegraphics[width=0.06\textwidth]{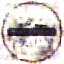}}\hfill
                \raisebox{-\height}{\includegraphics[width=0.06\textwidth]{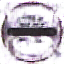}}\hfill
                \raisebox{-\height}{\includegraphics[width=0.06\textwidth]{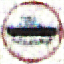}}\hfill
                \raisebox{-\height}{\includegraphics[width=0.06\textwidth]{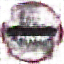}}\hfill
                \raisebox{-\height}{\includegraphics[width=0.06\textwidth]{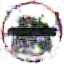}}\hfill
                \raisebox{-\height}{\includegraphics[width=0.06\textwidth]{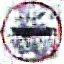}}\hfill
                \raisebox{-\height}{\includegraphics[width=0.06\textwidth]{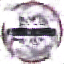}}\hfill                
            \end{subfigure}
            \caption{Images generated by pix2pix outside of training set (Zoll station)}
            \label{fig:oodZoll}
        \end{figure*}

\begin{figure*}[htb!] 
            \centering
            \begin{subfigure}{0.99\textwidth}
                \centering
                \raisebox{-\height}{\includegraphics[width=0.06\textwidth]{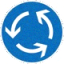}}
                \hfil
            \end{subfigure}
            \begin{subfigure}{0.99\textwidth}
                \raisebox{-\height}{\includegraphics[width=0.06\textwidth]{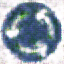}}\hfill
                \raisebox{-\height}{\includegraphics[width=0.06\textwidth]{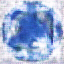}}\hfill
                \raisebox{-\height}{\includegraphics[width=0.06\textwidth]{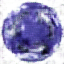}}\hfill
                \raisebox{-\height}{\includegraphics[width=0.06\textwidth]{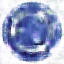}}\hfill
                \raisebox{-\height}{\includegraphics[width=0.06\textwidth]{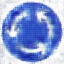}}\hfill
                \raisebox{-\height}{\includegraphics[width=0.06\textwidth]{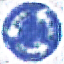}}\hfill
                \raisebox{-\height}{\includegraphics[width=0.06\textwidth]{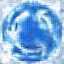}}\hfill
                \raisebox{-\height}{\includegraphics[width=0.06\textwidth]{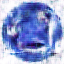}}\hfill
                \raisebox{-\height}{\includegraphics[width=0.06\textwidth]{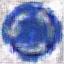}}\hfill
                \raisebox{-\height}{\includegraphics[width=0.06\textwidth]{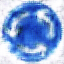}}\hfill                
                \raisebox{-\height}{\includegraphics[width=0.06\textwidth]{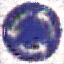}}\hfill
                \raisebox{-\height}{\includegraphics[width=0.06\textwidth]{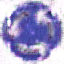}}\hfill
                \raisebox{-\height}{\includegraphics[width=0.06\textwidth]{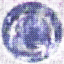}}\hfill
                \raisebox{-\height}{\includegraphics[width=0.06\textwidth]{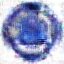}}\hfill
                \raisebox{-\height}{\includegraphics[width=0.06\textwidth]{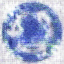}}\hfill
                
                \raisebox{-\height}{\includegraphics[width=0.06\textwidth]{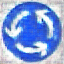}}\hfill
                \raisebox{-\height}{\includegraphics[width=0.06\textwidth]{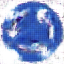}}\hfill
                \raisebox{-\height}{\includegraphics[width=0.06\textwidth]{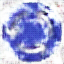}}\hfill
                \raisebox{-\height}{\includegraphics[width=0.06\textwidth]{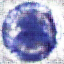}}\hfill
                \raisebox{-\height}{\includegraphics[width=0.06\textwidth]{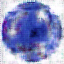}}\hfill                
                \raisebox{-\height}{\includegraphics[width=0.06\textwidth]{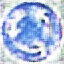}}\hfill
                \raisebox{-\height}{\includegraphics[width=0.06\textwidth]{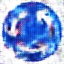}}\hfill
                \raisebox{-\height}{\includegraphics[width=0.06\textwidth]{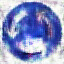}}\hfill
                \raisebox{-\height}{\includegraphics[width=0.06\textwidth]{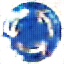}}\hfill
                \raisebox{-\height}{\includegraphics[width=0.06\textwidth]{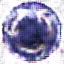}}\hfill
                \raisebox{-\height}{\includegraphics[width=0.06\textwidth]{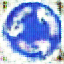}}\hfill
                \raisebox{-\height}{\includegraphics[width=0.06\textwidth]{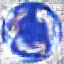}}\hfill
                \raisebox{-\height}{\includegraphics[width=0.06\textwidth]{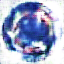}}\hfill
                \raisebox{-\height}{\includegraphics[width=0.06\textwidth]{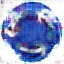}}\hfill
                \raisebox{-\height}{\includegraphics[width=0.06\textwidth]{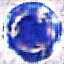}}\hfill
            \end{subfigure}
            \caption{Images generated by pix2pix outside of training set (Round about)}
            \label{fig:oodRoundAbout}
        \end{figure*}

\begin{figure*}[htb!]
            \centering
            \begin{subfigure}{0.99\textwidth}
                \centering
                \raisebox{-\height}{\includegraphics[width=0.06\textwidth]{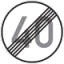}}
                \hfil
            \end{subfigure}
            \begin{subfigure}{0.99\textwidth}
                \raisebox{-\height}{\includegraphics[width=0.06\textwidth]{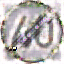}}\hfill
                \raisebox{-\height}{\includegraphics[width=0.06\textwidth]{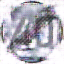}}\hfill
                \raisebox{-\height}{\includegraphics[width=0.06\textwidth]{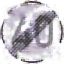}}\hfill
                \raisebox{-\height}{\includegraphics[width=0.06\textwidth]{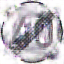}}\hfill
                \raisebox{-\height}{\includegraphics[width=0.06\textwidth]{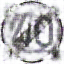}}\hfill
                \raisebox{-\height}{\includegraphics[width=0.06\textwidth]{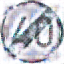}}\hfill
                \raisebox{-\height}{\includegraphics[width=0.06\textwidth]{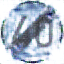}}\hfill
                \raisebox{-\height}{\includegraphics[width=0.06\textwidth]{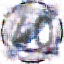}}\hfill
                \raisebox{-\height}{\includegraphics[width=0.06\textwidth]{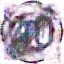}}\hfill
                \raisebox{-\height}{\includegraphics[width=0.06\textwidth]{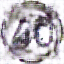}}\hfill                
                \raisebox{-\height}{\includegraphics[width=0.06\textwidth]{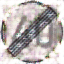}}\hfill
                \raisebox{-\height}{\includegraphics[width=0.06\textwidth]{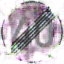}}\hfill
                \raisebox{-\height}{\includegraphics[width=0.06\textwidth]{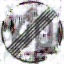}}\hfill
                \raisebox{-\height}{\includegraphics[width=0.06\textwidth]{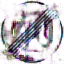}}\hfill
                \raisebox{-\height}{\includegraphics[width=0.06\textwidth]{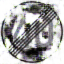}}\hfill
                
                \raisebox{-\height}{\includegraphics[width=0.06\textwidth]{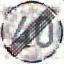}}\hfill
                \raisebox{-\height}{\includegraphics[width=0.06\textwidth]{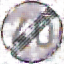}}\hfill
                \raisebox{-\height}{\includegraphics[width=0.06\textwidth]{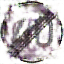}}\hfill
                \raisebox{-\height}{\includegraphics[width=0.06\textwidth]{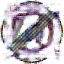}}\hfill
                \raisebox{-\height}{\includegraphics[width=0.06\textwidth]{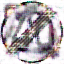}}\hfill                
                \raisebox{-\height}{\includegraphics[width=0.06\textwidth]{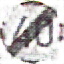}}\hfill
                \raisebox{-\height}{\includegraphics[width=0.06\textwidth]{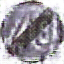}}\hfill
                \raisebox{-\height}{\includegraphics[width=0.06\textwidth]{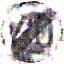}}\hfill
                \raisebox{-\height}{\includegraphics[width=0.06\textwidth]{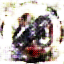}}\hfill
                \raisebox{-\height}{\includegraphics[width=0.06\textwidth]{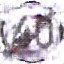}}\hfill
                \raisebox{-\height}{\includegraphics[width=0.06\textwidth]{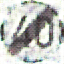}}\hfill
                \raisebox{-\height}{\includegraphics[width=0.06\textwidth]{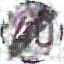}}\hfill
                \raisebox{-\height}{\includegraphics[width=0.06\textwidth]{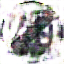}}\hfill
                \raisebox{-\height}{\includegraphics[width=0.06\textwidth]{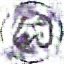}}\hfill
                \raisebox{-\height}{\includegraphics[width=0.06\textwidth]{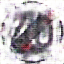}}\hfill
            \end{subfigure}
            \caption{Images generated by pix2pix outside of training set (end speed limit sign 40)}
            \label{fig:oodEndSpeed}
        \end{figure*}

To evaluate pix2pix generalization, we produced images from traffic signs that are not in the training set, shown in Figures \ref{fig:oodZoll}, \ref{fig:oodRoundAbout}, and \ref{fig:oodEndSpeed}. Visually inspecting these examples we see that the GAN does generalize, but it has trouble reproducing the text and digits in the signs, but overall it has no issue reproducing arrows.
    
\textbf{Baselines}. We compared the results with some traditional augmentation techniques: Blur, Brightness, Contrast, Displacement, Occlusion, Rotation, and Scaling. For each of these techniques defined each parameter in a range where the sign is still recognizable by a human, and then randomly sampled this space to generate new augmented examples with a uniform distribution.

\textbf{Model}. We use a CNN model based on SqueezeNet \cite{SqueezeNet}, motivated by its lightweight and computationally efficient architecture. Each combination of augmentation technique is trained five times, in order to account for variations in random weight initialization. We report the mean and standard deviation of accuracy across these runs. Each model is trained for 100 epochs using the Adam optimizer with a learning rate $\alpha = 0.01$.
        
        \begin{table} [htb!]
            \begin{center}
                \begin{tabular}{|l|c|c|c|c|} 
                    \hline                    
                    Augmentation & \# of Samples &\multicolumn{3}{c|}{ Accuracy (\%)}  \\
                    \cline{3-5}
                    & & $\mu \pm \sigma$ &Min &Max\\
                    \hline \hline
                    \begin{tabular}[c]{@{}c@{}c@{}}None\end{tabular} & 61089 & 92.1  $\pm$ 2.4 & 88.2  & 95.0   \\\hline
                    Blur  			& 122178 & 93.2 $\pm$ 1.3  & 91.4 &  94.7    \\\hline
                    Brightness  	& 122178 &94.9   $\pm$ 0.8 & 94.0 & 96.4    \\\hline
                    Contrast  		& 122178 & \textbf{95.5}  $\pm$ 0.8  & \textbf{94.6}  & \textbf{96.8}    \\\hline
                    Displacement  	& 122178 & 92.2  $\pm$ 1.8 & 89.2 & 94.2    \\\hline
                    Occlusion  		& 122178 & 93.9  $\pm$ 1.9 & 90.3 & 95.8   \\\hline
                    Rotation  		& 122178 & 94.6  $\pm$ 0.9 & 93.5 & 95.6     \\\hline
                    Scaling  		& 122178 & 94.8  $\pm$ 0.8 & 93.4 & 95.6   \\\hline
                    pix2pix 		& 122178 & 94.0 $\pm$ 1.0 & 92.2 & 95.0   \\\hline
                \end{tabular}
                \caption{Circular Traffic Signs - GAN augmentation compared to traditional augmentation regarding classification accuracy}
                \label{tab:results}
            \end{center}
        \end{table}
    
        \begin{table}[htb!]            
            \begin{center}
                \begin{tabular}{|l|c|c|c|c|} 
                    \hline
                    Augmentation & \# of Samples &\multicolumn{3}{c|}{ Accuracy (\%)}  \\
                    \cline{3-5}
                    & & $\mu \pm \sigma$ &Min &Max\\
                    \hline \hline                    
                    None		 	& 90218 & 93.8  $\pm$ 0.6 & 93.0  & 94.5 \\\hline
                    Blur  			& 180436 & 95.4 $\pm$ 0.5  & 95.0 &  96.3 \\\hline
                    Brightness  	& 180436 &96.2   $\pm$ 1.0 & 94.9 & 97.1 \\\hline
                    Contrast  		& 180436 &96.1  $\pm$ 0.5   & 95.6  & 96.6 \\\hline
                    Displacement  	& 180436 & \textbf{96.7  $\pm$ 0.4} & 96.3 & 97.2 \\\hline
                    Occlusion  		& 180436 & 95.3  $\pm$ 1.4 & 93.0 & 96.3 \\\hline
                    Rotation  		& 180436 & 95.4  $\pm$ 1.4 & 93.1 & 96.5   \\\hline
                    Scaling  		& 180436 & 94.8  $\pm$ 1.7 & 91.9 & 96.6 \\\hline
                    pix2pix 		& 180436 & 95.3 $\pm$ 0.3 & 95.0 & 95.8 \\\hline 
                    
                \end{tabular}	
                \caption {Triangular Traffic Signs - GAN augmentation compared to traditional augmentation regarding classification accuracy}
                \label{tab:resultsTri}
            \end{center}
        \end{table}

Results for circular traffic signs are presented in Table \ref{tab:results}. We see that augmentation with pix2pix increased accuracy by 2\% compared to the baseline without any augmentation. However we can also notice that mostly all other traditional augmentation techniques similarly produced increased accuracy. Our results show that the highest improvement was with contrast augmentation (95.5\%) and also the minimum and maximum accuracy were also the best. Results for triangular signs are shown in Table \ref{tab:resultsTri}, which follow a similar pattern of pix2pix augmented images producing accuracy improvements of around 2\%, but the classic augmentation technique of adding image displacements outperforms the baseline by 3\%.

        \begin{figure*}[htb!]
            \centering      
            \begin{subfigure}{0.32\textwidth}
                \includegraphics[width=\textwidth]{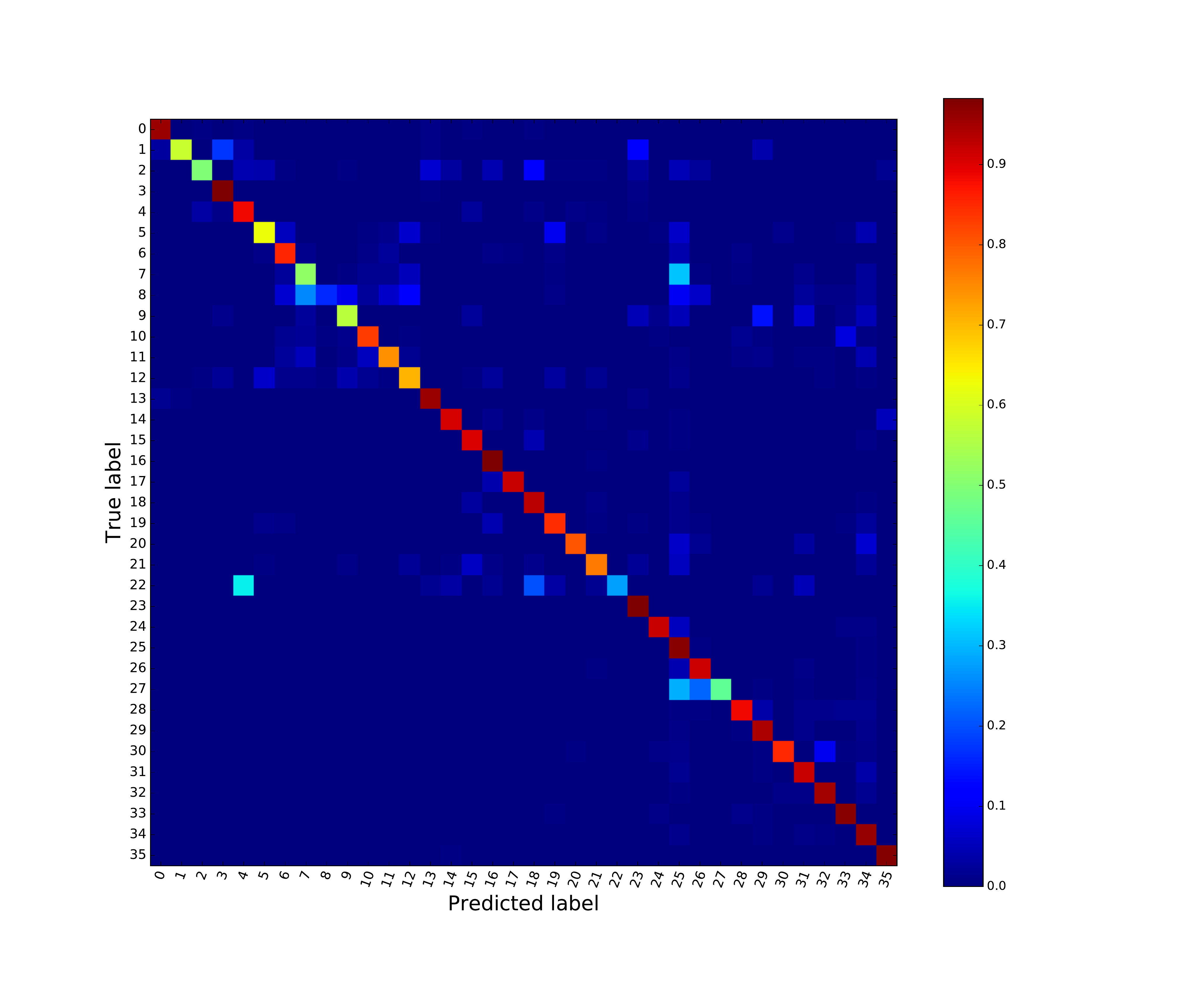}
                \caption{Baseline No Aug}
            \end{subfigure}      
            \begin{subfigure}{0.32\textwidth}
                \includegraphics[width=\textwidth]{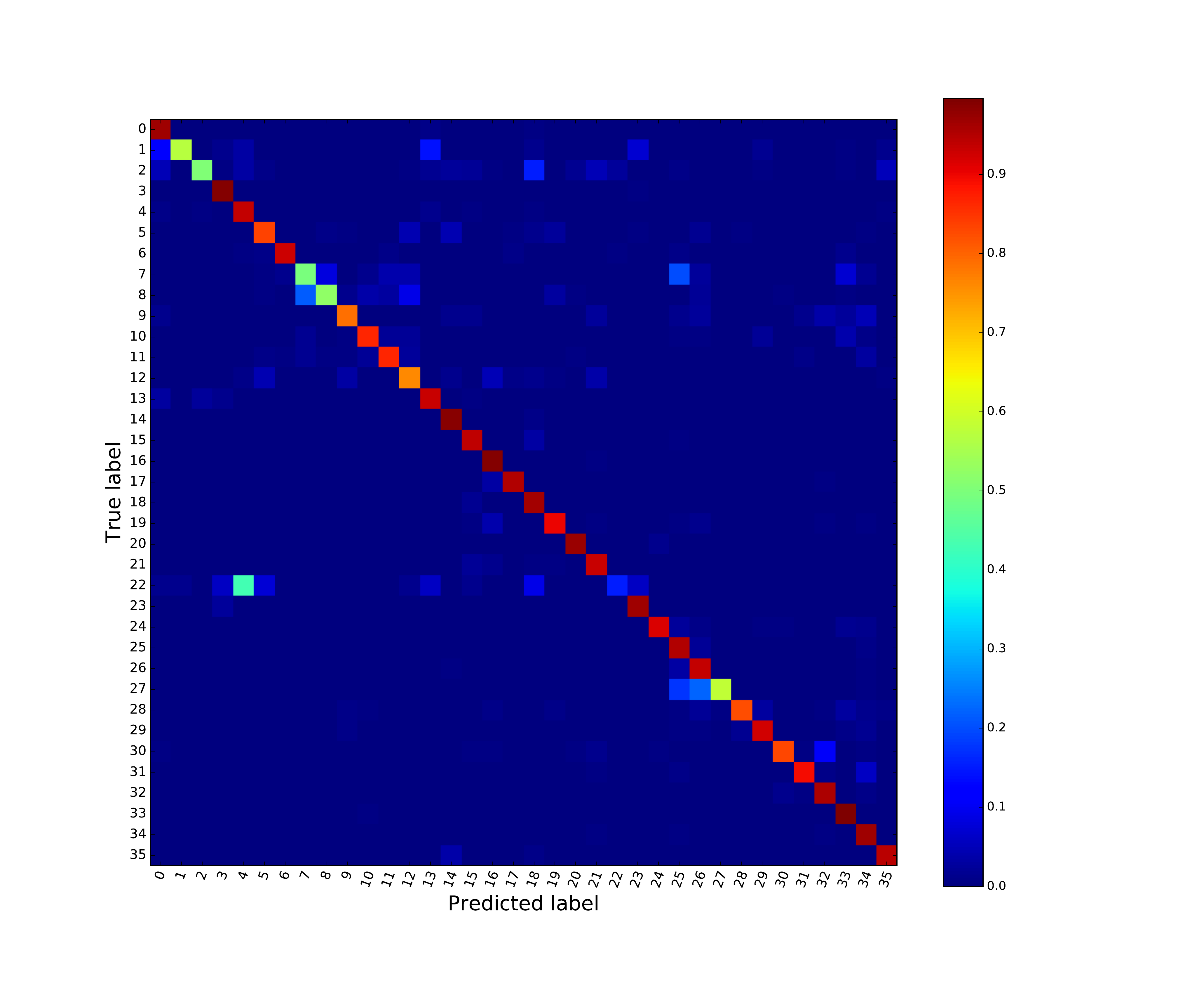}
                \caption{pix2pix Aug}
            \end{subfigure}      
            \begin{subfigure}{0.32\textwidth}
                \includegraphics[width=\textwidth]{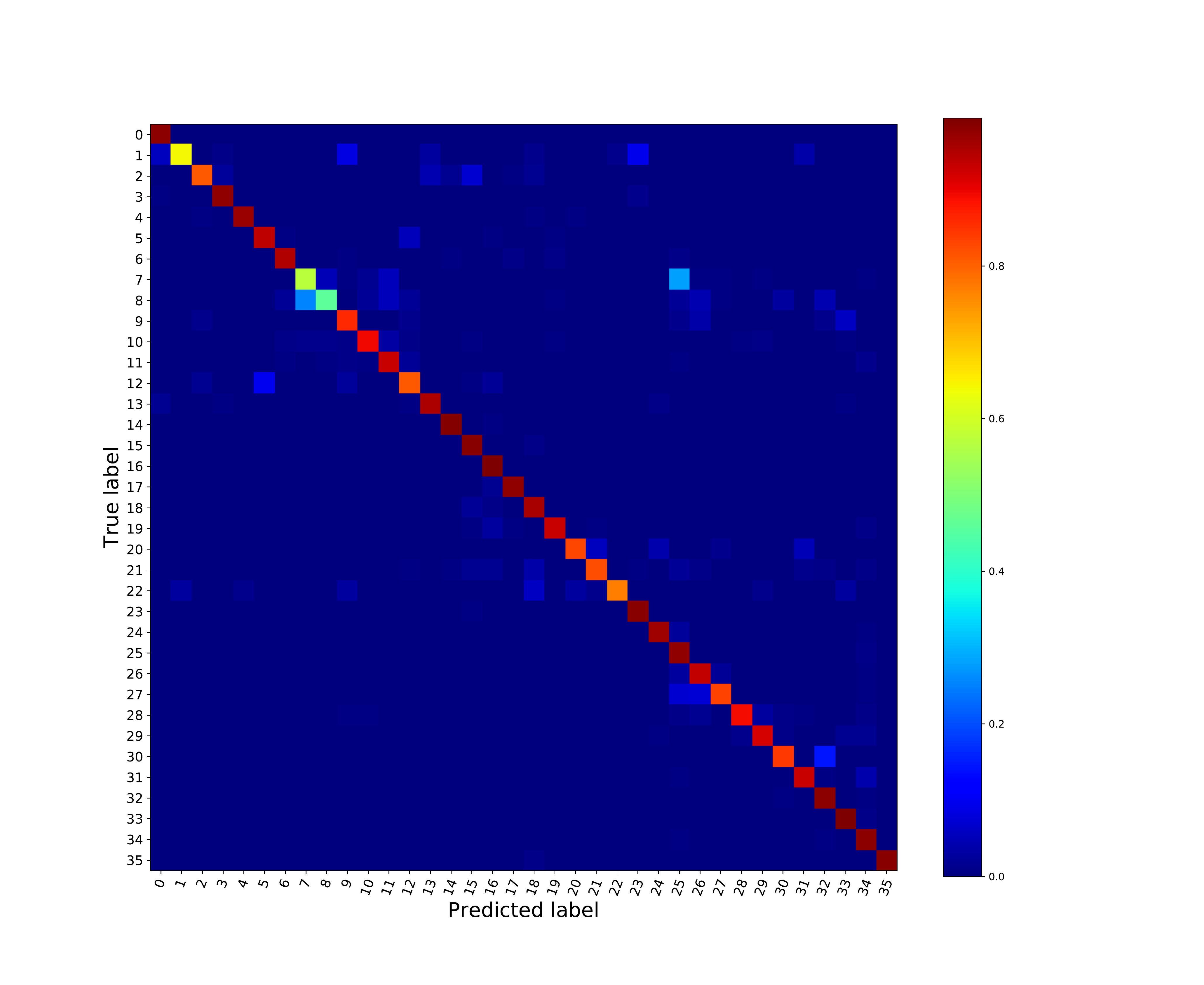}
                \caption{Contrast Aug}
            \end{subfigure}                  
            \caption{Comparison of Confusion Matrices for Circular Traffic Sign Classification}
            \label{fig:circularCM}
        \end{figure*}
        
\begin{figure*}[htb!]
            \centering      
            \begin{subfigure}{0.32\textwidth}
                \includegraphics[width=\textwidth]{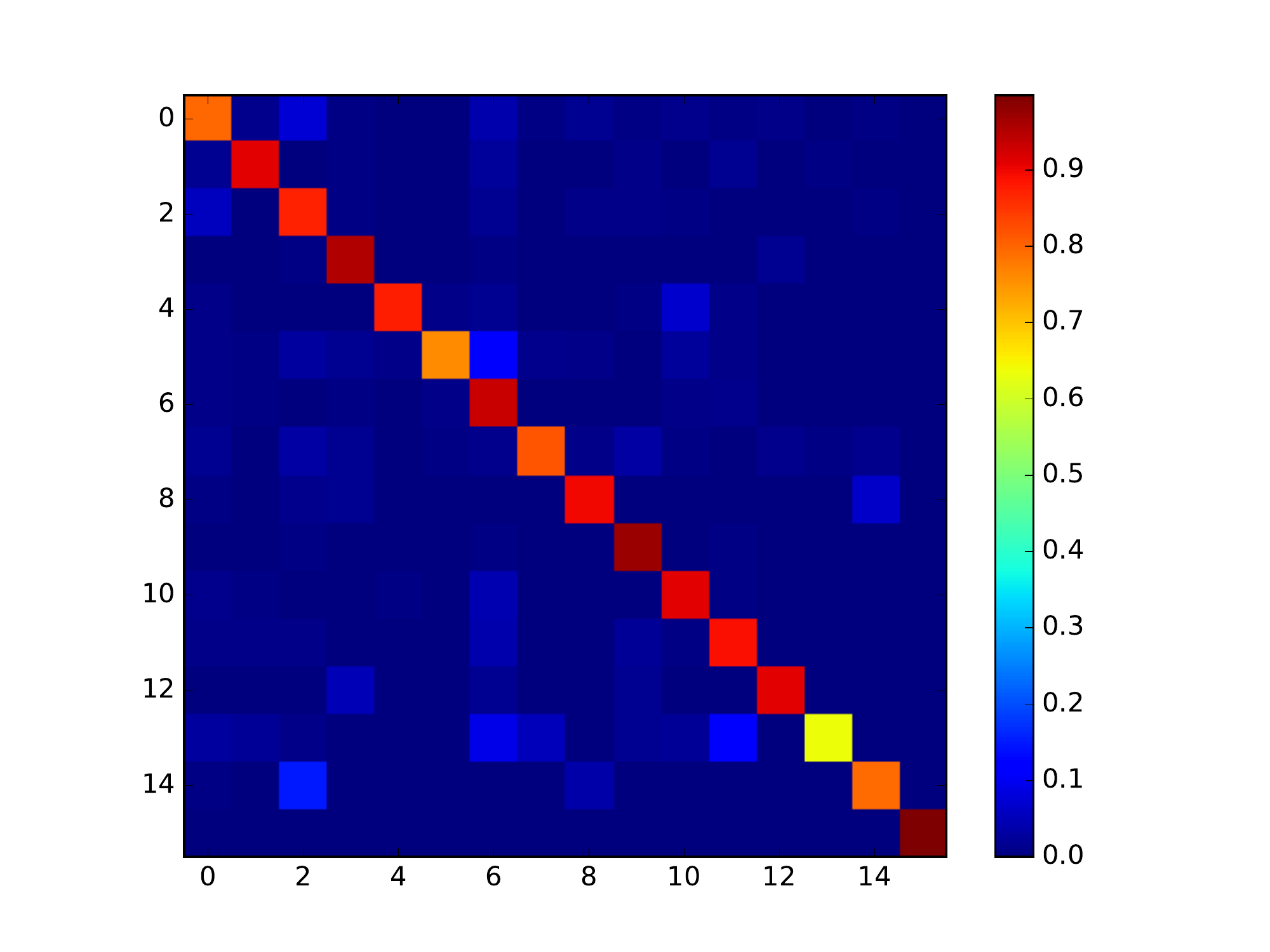}
                \caption{Baseline No Aug}
            \end{subfigure}      
            \begin{subfigure}{0.32\textwidth}
                \includegraphics[width=\textwidth]{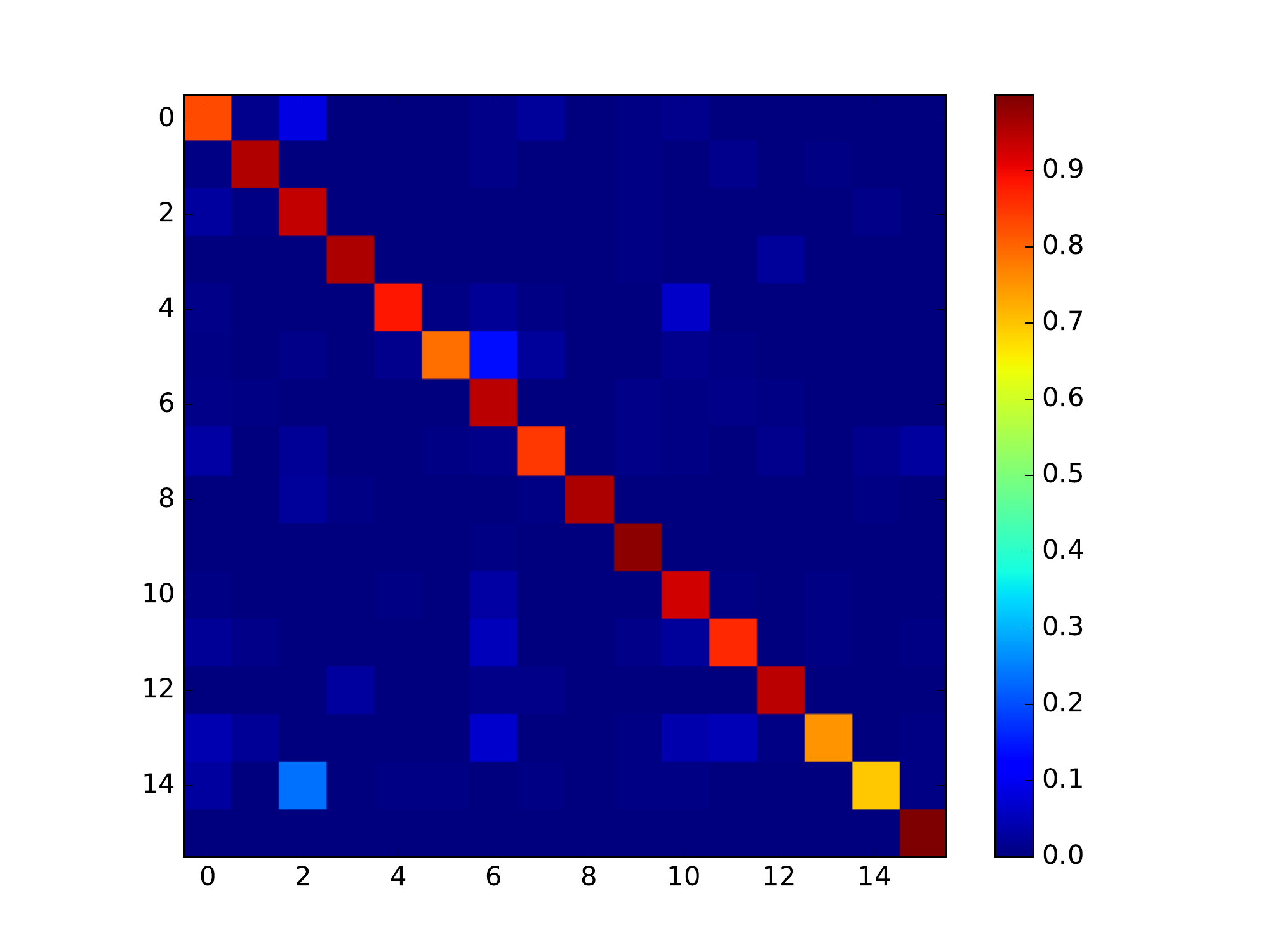}
                \caption{pix2pix Aug}
            \end{subfigure}      
            \begin{subfigure}{0.32\textwidth}
                \includegraphics[width=\textwidth]{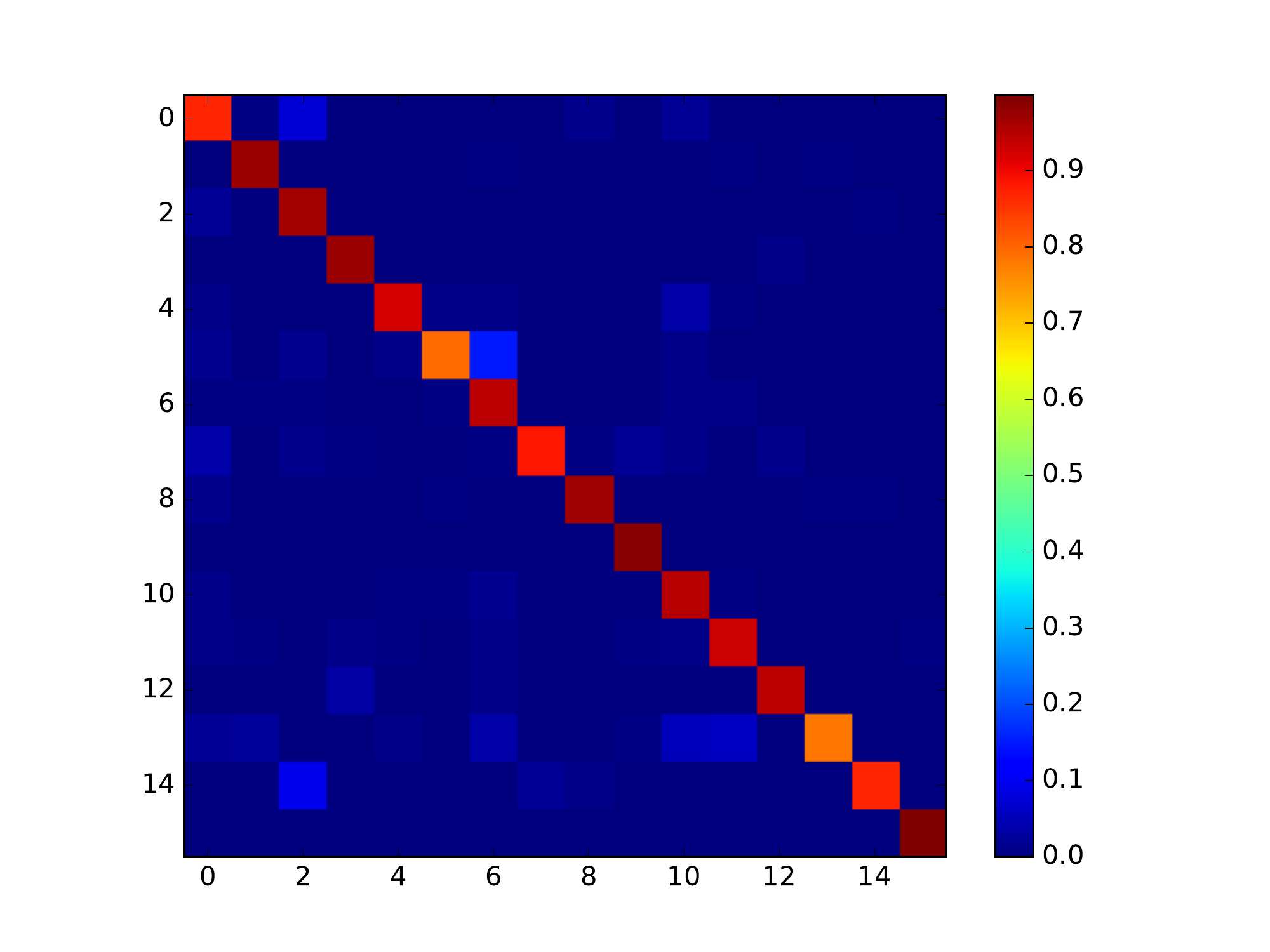}
                \caption{Contrast Aug}
            \end{subfigure}                  
            \caption{Comparison of Confusion Matrices for Triangular Traffic Sign Classification}
            \label{fig:triangularCM}
        \end{figure*}

Figures \ref{fig:circularCM} and \ref{fig:triangularCM} illustrate the confusion matrices of the baseline without any data augmentation, with pix2pix augmentation, and with the best performing augmentation respectively.
        
We can see that our circular baseline has a lot of miss-classifications, for example there was a confusion with class 8 (end speed limit 70) to be classified as class 6 (end no passing trucks), 7 (end speed limit 30) or 9 (end speed limit 80), this confusion was reduced by using pix2pix augmentation and with each of blur, brightness, contrast, occlusion, rotation and scaling augmentation, even though there is still some miss-classification between class 8 and 7.

Another conflict we notice in baseline confusion matrix is of class 22 (pass left) to be classified as class 4 (direction right), which might have happened because they both sign are in blue color and there were less training samples of class 22 in the original dataset. This confusion disappeared using contrast, rotation and scaling augmentation, and reduced using blur augmentation. But it remained with pix2pix augmentation.

Moreover we can see that in baseline confusion matrix there is a miss-classification of class 27 (speed limit 40) with both classes 26 (speed limit 30) and 25 (speed limit 20), which is reasonable due to the similarity in colors and shapes of these signs. Nevertheless it was eliminated with each of brightness and contrast, but it did not improve with pix2pix augmentation.
        
For triangular traffic signs (Figure \ref{fig:triangularCM}), we see confusion increment for Traffic Jam (Class 11) and Two-way Traffic (Class 14), and confusion decreases with Lane merge (Class 8), Bicycle lane (Class 0), and Pedestrians (Class 2). Overall the bigggest improvement of traditional augmentation is produced in triangular traffic signs, clearly outperforming a pix2pix augmentation.

        \begin{table*}[htb!]            
            \begin{center}                
                \begin{tabular}{|c|c|c|c|c|c|c|c|c|}                     
                    \hline
                    \multicolumn{2}{|c|}{Conv Layers}
                    &\begin{tabular}[c]{@{}c@{}c@{}}\\Training \\ Samples \end{tabular}&\multicolumn{3}{c|}{Accuracy (\%)} & \multicolumn{3}{c|}{Balanced Accuracy (\%)}  \\
                    \cline{1-2}\cline{4-9}
                    Discriminator &Generator& &($\mu \pm \sigma$)&Min &Max & ($\mu \pm \sigma$)&Min &Max\\ 
                    \hline \hline
                    \multicolumn{2}{|c|}{No augmentation} & 61089 & 92.1  $\pm$ 2.4 & 88.2  & 95.0 & 80.3 $\pm$ 3.7 & 75.4 & 86.1  \\\hline
                    3&	2 & 66898 & 90.7 $\pm$ 1.6 & 87.9 & 92.7  & 80.8  $\pm$ 1.2 & 79.6 & 82.6  \\\hline
                    3&	4 & 66898 & 91.8 $\pm$ 3.8 & 84.4 & 94.6  & 82.4 $\pm$ 2.6 & 77.8 & 85.9 \\\hline
                    3&	6 & 66898 & 91.7 $\pm$ 1.6 & 89.2 & 93.4  & 79.4 $\pm$ 2.9 & 75.1 & 84.2 \\\hline
                    3&	8 & 66898 & 92.2 $\pm$ 2.5 & 88.9 & 94.5  & 79.6  $\pm$ 5.5  & 70.4 & 84.3 \\\hline
                    3&	10 & 66898 & 91.1  $\pm$ 2.5 & 86.9 & 94.6  & 81.8 $\pm$ 1.8 & 79.5 & 84.4  \\\hline
                    4 &2 & 66898 & 92.7 $\pm$ 0.8 & 91.4 & 93.9  & 81.7  $\pm$ 2.1 & 78.4 & 84.4 \\\hline
                    
                    4 &4 & 66898 & \textbf{93.9 $\pm$ 0.6} & 93.0 & 94.8 & \textbf{ 83.6 $\pm$ 1.6} & 81.4 & 86.0 \\\hline
                    4 &6 & 66898 & 92.6 $\pm$ 1.6 & 89.7 & 94.0  & 80.6 $\pm$ 1.2 & 79.0 & 81.8 \\\hline
                    4 &8 & 66898 & 92.6 $\pm$ 1.1 & 91.2 & 94.3  & 80.3  $\pm$ 1.0  & 79.0 & 81.9 \\\hline
                    4 &10 & 66898 & 91.8  $\pm$ 2.1 & 88.5 & 95.3  & 82.0 $\pm$ 2.9 & 78.5 & 85.0  \\\hline
                \end{tabular}	
                \caption{Hyper-parameter Tuning of the pix2pix GAN Architecture on Circular Traffic Signs}
                \label{tab:accuracyCircular} 
            \end{center}
        \end{table*}
        
        \begin{table*}[htb!]
            \begin{center}
                \begin{tabular}{|c|c|c|c|c|c|c|c|c|} 
                    \hline
                    \multicolumn{2}{|c|}{Conv Layers}
                    &\begin{tabular}[c]{@{}c@{}c@{}}\\Training \\ Samples \end{tabular}&\multicolumn{3}{c|}{Accuracy (\%)} & \multicolumn{3}{c|}{Balanced Accuracy (\%)}  \\
                    \cline{1-2}\cline{4-9}
                    Discriminator &Generator& &($\mu \pm \sigma$)&Min &Max & ($\mu \pm \sigma$)&Min &Max\\ 
                    \hline \hline
                    \multicolumn{2}{|c|}{No augmentation} & 61089 & 92.1  $\pm$ 2.4 & 88.2  & 95.0 & 80.3 $\pm$ 3.7 & 75.4 & 86.1  \\\hline
                    3&	2  & 66898 & 92.2 $\pm$ 1.2 & 90.1 & 93.3  & 81.9  $\pm$ 2.5 & 78.8 & 85.3 \\\hline
                    3&	4  & 66898 & 92.3 $\pm$ 1.5 & 90.7 & 94.8   & 80.6 $\pm$ 3.0 & 76.6 & 85.3 \\\hline
                    3&	6  & 66898 & 92.5 $\pm$ 1.2 & 90.3 & 93.4  & 80.8 $\pm$ 2.8 & 77.4 & 83.7 \\\hline
                    3&	8  & 66898 & 86.8 $\pm$ 0.9 & 91.5 & 94.0  & 79.7  $\pm$ 2.5  & 76.2 & 82.8 \\\hline
                    3&	10 & 66898 & 93.2  $\pm$ 1.0 & 91.9 & 94.4  & 81.9 $\pm$ 1.8 & 78.3 & 83.6  \\\hline
                    
                    4&	2  & 66898 & 93.7 $\pm$ 1.0 & 92.2 & 94.9  & 83.1  $\pm$ 0.7 & 82.3 & 83.8 \\\hline
                    4&	4  & 66898 & 93.2 $\pm$ 1.0 & 91.5 & 94.2   & 83.1 $\pm$ 1.1 & 81.5 & 84.3 \\\hline
                    
                    4&	6  & 66898 & \textbf{94.4 $\pm$ 0.7} & 93.5 & 95.3 & \textbf{84.5 $\pm$ 2.1} & 81.6 & 87.8 \\\hline
                    4&	8  & 66898 & 91.5 $\pm$ 1.9 & 90.3 & 94.8  & 81.6  $\pm$ 1.4  & 79.8 & 83.4 \\\hline
                    4&	10 & 66898 & 93.7  $\pm$ 1.2 & 91.6 & 95.2  & 83.7 $\pm$ 1.9 & 81.9 & 87.2 \\\hline
                \end{tabular}	
                \caption {Hyper-parameter Tuning of the pix2pix GAN Architecture on Triangular Traffic Signs}
                \label{tab:accuracyTriangular} 
            \end{center}
        \end{table*}
        
\section{Conclusions and Future Work}
\label{sec:conclusion}
Data augmentation is essential in every classification problem, specially when we use CNNs. In this work we studied and evaluated the effects of using GANs for data augmentation in traffic sign classification. We used pix2pix to generate traffic sign images from symbolic ones, as an alternative to mode collapse in a Conditional GAN, which might be more appropriate for this kind of problem.
        
With this GAN architecture, augmented images cause a circular traffic sign classification accuracy increase from 92.1\% to 94.0\%. But applying traditional augmentation techniques was more successful, which can reach 95.5\% with contrast augmentation. For triangular traffic signs, we observe an increase from baseline 93.8\% to 95.3\%, but again this is outperformed by a classic augmentation technique of adding displacement, increasing accuracy to 96.7\%.
        
Designing data augmentation techniques is difficult, and GANs still seem promising in automating this process. For future work we wish to evaluate a type of GAN that was specifically designed for data augmentation  \cite{antoniou2017data}. Overall we expected that GANs produced a bigger improvement, but our negative result shows that still there is much research needed in terms of variability in the datasets and how a GAN can generate novel images from it.
        
\section*{Acknowledgements}
        
Nour Soufi gratefully acknowledges the support given by the Autonomous Systems Group and the Institute of Visual Computing  at the Hochschule Bonn-Rhein-Sieg, and the B-IT Bonn-Aachen International Center for Information Technology.

Matias Valdenegro-Toro was partially supported by the FP7-PEOPLE-2013-ITN project ROBOCADEMY (Ref 608096) funded by the European Commission.

\bibliographystyle{IEEEtran}
\bibliography{bibliography}

\begin{thebibliography}{10}
\providecommand{\url}[1]{#1}
\csname url@samestyle\endcsname
\providecommand{\newblock}{\relax}
\providecommand{\bibinfo}[2]{#2}
\providecommand{\BIBentrySTDinterwordspacing}{\spaceskip=0pt\relax}
\providecommand{\BIBentryALTinterwordstretchfactor}{4}
\providecommand{\BIBentryALTinterwordspacing}{\spaceskip=\fontdimen2\font plus
\BIBentryALTinterwordstretchfactor\fontdimen3\font minus
  \fontdimen4\font\relax}
\providecommand{\BIBforeignlanguage}[2]{{%
\expandafter\ifx\csname l@#1\endcsname\relax
\typeout{** WARNING: IEEEtran.bst: No hyphenation pattern has been}%
\typeout{** loaded for the language `#1'. Using the pattern for}%
\typeout{** the default language instead.}%
\else
\language=\csname l@#1\endcsname
\fi
#2}}
\providecommand{\BIBdecl}{\relax}
\BIBdecl

\bibitem{SqueezeNet}
F.~Iandola, S.~Han, M.~W.~Moskewicz, K.~Ashraf, W.~Dally, and K.~Keutzer,
  ``Squeezenet: Alexnet-level accuracy with 50x fewer parameters and <0.5mb
  model size,'' 02 2016.

\bibitem{pix2pix2016}
P.~Isola, J.-Y. Zhu, T.~Zhou, and A.~A. Efros, ``Image-to-image translation
  with conditional adversarial networks,'' \emph{arxiv}, 2016.

\bibitem{7605450}
Y.~Yuan, Z.~Xiong, and Q.~Wang, ``An incremental framework for video-based
  traffic sign detection, tracking, and recognition,'' \emph{IEEE Transactions
  on Intelligent Transportation Systems}, vol.~18, no.~7, pp. 1918--1929, 2017.

\bibitem{8001185}
E.~Okafor, R.~Smit, L.~Schomaker, and M.~Wiering, ``Operational data
  augmentation in classifying single aerial images of animals,'' in
  \emph{INnovations in Intelligent SysTems and Applications (INISTA), 2017 IEEE
  International Conference on}.\hskip 1em plus 0.5em minus 0.4em\relax IEEE,
  2017, pp. 354--360.

\bibitem{6033589}
P.~Sermanet and Y.~LeCun, ``Traffic sign recognition with multi-scale
  convolutional networks,'' in \emph{Neural Networks (IJCNN), The 2011
  International Joint Conference on}.\hskip 1em plus 0.5em minus 0.4em\relax
  IEEE, 2011, pp. 2809--2813.

\bibitem{6033458}
D.~Cire{\c{s}}an, U.~Meier, J.~Masci, and J.~Schmidhuber, ``A committee of
  neural networks for traffic sign classification,'' in \emph{Neural Networks
  (IJCNN), The 2011 International Joint Conference on}.\hskip 1em plus 0.5em
  minus 0.4em\relax IEEE, 2011, pp. 1918--1921.

\bibitem{6033395}
J.~Stallkamp, M.~Schlipsing, J.~Salmen, and C.~Igel, ``The german traffic sign
  recognition benchmark: a multi-class classification competition,'' in
  \emph{Neural Networks (IJCNN), The 2011 International Joint Conference
  on}.\hskip 1em plus 0.5em minus 0.4em\relax IEEE, 2011, pp. 1453--1460.

\bibitem{6033494}
F.~Zaklouta, B.~Stanciulescu, and O.~Hamdoun, ``Traffic sign classification
  using kd trees and random forests,'' in \emph{Neural Networks (IJCNN), The
  2011 International Joint Conference on}.\hskip 1em plus 0.5em minus
  0.4em\relax IEEE, 2011, pp. 2151--2155.

\bibitem{6766231}
J.~Jin, K.~Fu, and C.~Zhang, ``Traffic sign recognition with hinge loss trained
  convolutional neural networks,'' \emph{IEEE Transactions on Intelligent
  Transportation Systems}, vol.~15, no.~5, pp. 1991--2000, 2014.

\bibitem{8279326}
L.~Wen and K.-H. Jo, ``Traffic sign recognition and classification with
  modified residual networks,'' in \emph{System Integration (SII), 2017
  IEEE/SICE International Symposium on}.\hskip 1em plus 0.5em minus 0.4em\relax
  IEEE, 2017, pp. 835--840.

\bibitem{ashiquzzaman2017applying}
A.~Ashiquzzaman, A.~K. Tushar, and A.~Rahman, ``Applying data augmentation to
  handwritten arabic numeral recognition using deep learning neural networks,''
  \emph{arXiv preprint arXiv:1708.05969}, 2017.

\bibitem{NIPS2012_4824}
\BIBentryALTinterwordspacing
A.~Krizhevsky, I.~Sutskever, and G.~E. Hinton, ``Imagenet classification with
  deep convolutional neural networks,'' in \emph{Advances in Neural Information
  Processing Systems 25}, F.~Pereira, C.~J.~C. Burges, L.~Bottou, and K.~Q.
  Weinberger, Eds.\hskip 1em plus 0.5em minus 0.4em\relax Curran Associates,
  Inc., 2012, pp. 1097--1105. [Online]. Available:
  \url{http://papers.nips.cc/paper/4824-imagenet-classification-with-deep-convolutional-neural-networks.pdf}
\BIBentrySTDinterwordspacing

\bibitem{wu2015deep}
R.~Wu, S.~Yan, Y.~Shan, Q.~Dang, and G.~Sun, ``Deep image: Scaling up image
  recognition,'' \emph{arXiv preprint arXiv:1501.02876}, vol.~7, no.~8, 2015.

\bibitem{NIPS2014_5423}
\BIBentryALTinterwordspacing
I.~Goodfellow, J.~Pouget-Abadie, M.~Mirza, B.~Xu, D.~Warde-Farley, S.~Ozair,
  A.~Courville, and Y.~Bengio, ``{Generative Adversarial Nets},'' in
  \emph{{Advances in Neural Information Processing Systems 27}}, Z.~Ghahramani,
  M.~Welling, C.~Cortes, N.~D. Lawrence, and K.~Q. Weinberger, Eds.\hskip 1em
  plus 0.5em minus 0.4em\relax Curran Associates, Inc., 2014, pp. 2672--2680.
  [Online]. Available:
  \url{http://papers.nips.cc/paper/5423-generative-adversarial-nets.pdf}
\BIBentrySTDinterwordspacing

\bibitem{mirza2014conditional}
M.~Mirza and S.~Osindero, ``Conditional generative adversarial nets,''
  \emph{arXiv preprint arXiv:1411.1784}, 2014.

\bibitem{antoniou2017data}
A.~Antoniou, A.~Storkey, and H.~Edwards, ``Data augmentation generative
  adversarial networks,'' \emph{arXiv preprint arXiv:1711.04340}, 2017.

\end{thebibliography}

\end{document}